%% file: egpaper_for_review.tex
\documentclass[10pt,twocolumn,letterpaper]{article}

\usepackage{cvpr}
\usepackage{times}
\usepackage{epsfig}
\usepackage{graphicx}
\usepackage{amsmath}
\usepackage{amssymb}
\usepackage{sidecap}
\usepackage{xcolor}
\usepackage{pbox}

\usepackage{multirow}
\usepackage{booktabs}

\usepackage[pagebackref=true,breaklinks=true,letterpaper=true,colorlinks,bookmarks=false]{hyperref}

\cvprfinalcopy 

\def\cvprPaperID{****} 

\ifcvprfinal\fi
\begin{document}

\newcommand{\welambdaten}{$\text{WE}_{\lambda=10}$}
\newcommand{\welambdazero}{$\text{WE}_{\lambda=0}$}
\newcommand{\xv}{\mathbf{x}^v}
\newcommand{\xl}{\mathbf{x}^l}

\title{Unifying Few- and Zero-Shot Egocentric Action Recognition}

\author{Tyler R.~Scott\thanks{Research conducted during an internship at Facebook Reality Labs.}\\
University of Colorado, Boulder\\
{\tt\small tysc7237@colorado.edu}
\and
Michael Shvartsman\\
Facebook Reality Labs\\
{\tt\small michael.shvartsman@fb.com}
\and
Karl Ridgeway\\
Facebook Reality Labs\\
{\tt\small karl.ridgeway@fb.com}
}

\maketitle

\begin{abstract}
Although there has been significant research in egocentric action recognition, most methods and tasks, including EPIC-KITCHENS, suppose a fixed set of action classes.
Fixed-set classification is useful for benchmarking methods, but is often unrealistic in practical settings due to the compositionality of actions, resulting in a functionally infinite-cardinality label set.
In this work, we explore generalization with an open set of classes by unifying two popular approaches: few- and zero-shot generalization (the latter which we reframe as cross-modal few-shot generalization). 
We propose a new set of splits derived from the EPIC-KITCHENS dataset that allow evaluation of open-set classification, and use these splits to show 
that adding a metric-learning loss to the conventional direct-alignment baseline can improve zero-shot classification by as much as 10\%, while not 
sacrificing few-shot performance.
\end{abstract}

\section{Introduction}

The egocentric action recognition task consists of observing short first-person video segments of an action being performed,
and predicting the label---typically a verb--noun pair---that a human would assign (e.g., `pick-up plate', or `mix pasta').
Many supervised models (e.g., \cite{carreira2017quo, zhou2018temporal}) treat the problem as a \emph{fixed-set} classification task, where the set of action classes is identical during training and evaluation.

A model trained with the traditional fixed-set approach is encouraged to output an orthogonal basis over classes, which (1) requires a pre-determined number of outputs, preventing the prediction of unseen classes (i.e., novel verbs and nouns) and (2) conceals the semantic structure of actions (e.g., the verb `take' is more similar to `put' than `mix') in intermediate representations of the model.
We address both issues. We treat egocentric action recognition as an \emph{open-set} generalization task, where model performance is reported 
on held-out classes, and we utilize metric-learning losses, one approach for capturing the semantic structure of the label space.

We consider two popular paradigms for open-set evaluation: few-shot generalization (FSG; \cite{Snell2017}) and zero-shot-generalization (ZSG; \cite{Socher2013}). In the former, we use the model to classify \emph{query} instances from classes unseen during training using a small \emph{support set} of labeled samples. In the latter, we use the model to map videos to a latent representation that captures the semantic structure of the label space, and recognize instances from new classes by matching them to prototypes that are known a priori. 

While the two have been proposed as separate tasks, we recognize that ZSG can be framed as another instance of FSG, in which the support 
set contains a semantic representation of the class labels. We use this insight to generalize ZSG to a task we term \emph{cross-modal few-shot generalization} (CM-FSG). CM-FSG includes ZSG, as well as other task variants such as ones where the cross-modal information is not derived from language, or multiple instances of the semantic representation are available for a class. 

In this work, we identify four main contributions:
first, we formally unify FSG and CM-FSG into a framework that promotes inter-method comparison and provides the ability to compare open-set tasks.
Second, we present three new data splits from the original EPIC-KITCHENS training set---each with its own train, validation, and test subset---specifically designed to evaluate open-set generalization. 
Third, we explore several candidate loss functions to train neural networks to jointly perform the two tasks.
Fourth, we conduct a head-to-head comparison of FSG and CM-FSG on identical data splits and show that among the methods explored, the ones that do best in one task also do best in the other (i.e., there is no performance trade-off).
In addition, our results emphasize the importance of metric-learning losses not only for FSG, but CM-FSG, where we observe improvements upwards of 10\% over the conventional baseline.
We hope our work bridges advancements in open-set classification with egocentric action recognition, and that our results serve as a first benchmark.

\section{Open-Set Generalization Tasks}

Below we formalize the two open-set generalization tasks with respect to action recognition. 
Let  $\mathbf{x}^v \in \mathbb{R}^{F \times C \times H \times W}$ denote an input video clip consisting of $F$ $C$-channel frames with height, $H$, and width, $W$, and let $x^l$ denote an action label (e.g., `take fork').
Both FSG and ZSG are evaluated \emph{episodically}, where each episode contains a random sample of $n$ action classes, denoted $\mathcal{Y}$, which are disjoint from the set of training action classes, $\mathcal{Y}_{\text{train}}$ (i.e., $\mathcal{Y} \cap \mathcal{Y}_{\text{train}} = \emptyset$).
We make the distinction between the action class (e.g., `class 345') and the semantic action label (e.g., `take fork') explicit here, as this is what allows us to unify FSG and ZSG in a common framing below. 

\subsection{Few-Shot}

In FSG, the goal is to generalize to classes in $\mathcal{Y}$ using only a few video instances from each.
In each episode, $k + m$ instances are sampled from each class in $\mathcal{Y}$. The first $k$ instances (or `shots' from `few-shot') make up the \emph{support set},
$\mathcal{S}$, and the remaining $m$ make up the \emph{query set}, $\mathcal{Q}$:
\begin{equation}
\begin{split}
&\mathcal{S} = \{(\xv_{ij}, y_{ij}) \vert y_{ij} \in \mathcal{Y}\}_{i=1:n, \; j=1:k}, \\
&\mathcal{Q} = \{\xv_{ij}\}_{i=1:n, \; j=k+1:k+m}, 
\end{split}
\end{equation}
where $\xv_{ij}$ is the $j$th video instance of the $i$th class in the episode. 
Evaluation proceeds by classifying each element in the query set using the support set.
In EPIC-KITCHENS, there are a number of classes with very few instances. 
Therefore, during evaluation, we sample up to $m$ query instances per class. 
Since every episode will have a different number of queries, we report accuracy over all episodes.

\subsection{Cross-Modal Few-Shot}

To see how FSG is related to ZSG, recall above the distinction we made between the set of action labels and the set of action classes. 
The action labels are ignored in standard FSG, since each support tuple consists of a video and class. 
If one were to replace the video, $\mathbf{x}^v$, with the natural language description of the action class (e.g., `take cup', denoted by $\mathbf{x}^l$), one would be in a \emph{cross-modal few-shot generalization} (CM-FSG) setting, where the support set contains the action labels associated with each of the $n$ classes in $\mathcal{Y}$, and the query set remains unchanged:
\begin{equation}
\begin{split}
\mathcal{S}& = \{(\xl_{ij}, y_{ij}) \vert y_{ij} \in \mathcal{Y}\}_{i=1:n, j=1:k}, \\
\mathcal{Q}& = \{\xv_{ij}\}_{i=1:n, \; j=k+1:k+m}.
\end{split}
\end{equation}

When $k=1$, CM-FSG reduces to ZSG. When $k > 1$, we obtain a novel task. This novel task is not possible in the conventional ZSG setting since the distinct instances of a class are identical (i.e., they are all the same action label), but can be possible with noisy labels or the richer narrations from which the label is generated. 

\section{Related Work}

We now highlight several common approaches for FSG and ZSG.
Many FSG methods learn an \emph{embedding} of the inputs---typically images or video clips---where samples that are farther apart are less likely
to be from the same class. These methods typically make use of pairwise \cite{Hadsell2006}, triplet \cite{Schroff2015,Wang2019},
quadruplet \cite{Ustinova2016}, or group-based \cite{Snell2017} constraints via metric-learning loss functions, to promote intra-class similarity and inter-class dissimilarity.
Another popular approach is meta-learning \cite{Finn2017}, which is focused on learning to quickly adapt models to unseen classes. 
Memory-augmented neural networks \cite{Santoro2016b} have also been explored because they can use external memory mechanisms to store and recall data from unseen classes.
Recently, the above few-shot methods have begun being applied in the domain of action recognition \cite{Bishay2019,Cao2019,Careaga2019,Mishra2018,Zhu2018}.

For ZSG, there are three common approaches: (1) learn a function that maps inputs directly to an attribute vector,
where new classes constitute novel compositions of attributes \cite{Palatucci2009}, (2) map inputs into a pretrained semantic space (e.g., Word2Vec or BERT), where new classes can be directly interpreted \cite{Hahn2019,Socher2013}, and (3) 
learn two functions that map inputs and attribute/semantic vectors, respectively, to a joint latent space \cite{Bishay2019,Mishra2018,Snell2017}.
In approaches (1) and (2), the desired representation is typically fixed---either predefined class-attribute vectors
or predefined word embeddings. This discourages the model from representing features that are unique to
the input space, in our case, the visual and temporal features from video that may not correspond to
semantic features of the labels (e.g., that bananas tend to be yellow). 
These approaches are similarly applicable to CM-FSG.
We explore metric-learning methods from FSG in conjunction with approaches (2) and (3) from ZSG further in Section \ref{sec:models}.

\section{Methods}
\label{sec:models}

\begin{figure*}[htbp]
\centering
\includegraphics[width=\textwidth]{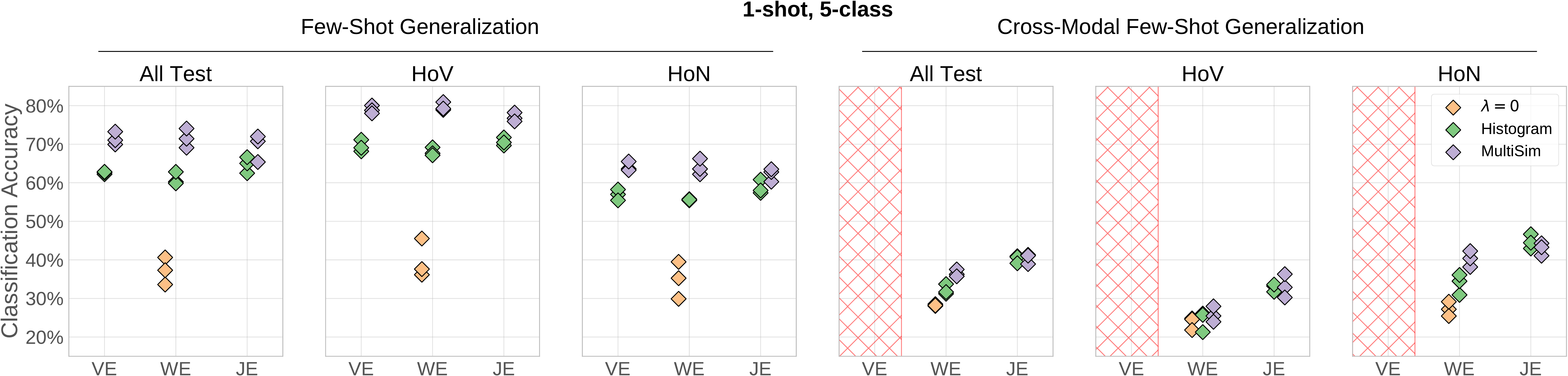}
\caption{Classification accuracy, computed over 500 test episodes, for the video embedding (VE), word embedding (WE), and joint embedding (JE). Both FSG and CM-FSG are evaluated using 1 shot, 20 queries, and 5 classes per episode (i.e., $k = 1$, $m = 20$, and $n = 5$). Each pane is characterized according to a generalization task (FSG or CM-FSG) and a subset of the test set (All Test, HoV, or HoN). For a given generalization task, test subset, and method, the same-colored points represent performance on each of the three data splits. The red hatching indicates that VE cannot be evaluated using CM-FSG.}
\label{fig:results}
\end{figure*}

Generalizing FSG and CM-FSG into a common framework lets us seek a method that is capable of performing successfully in both tasks. We begin by introducing a video embedding (a unimodal FSG-only method) and then extend it to methods that perform both tasks. 

\paragraph{Video Embedding (VE)} 
VE learns a deep embedding using video instances, similar to \cite{Careaga2019}. This is an FSG-only baseline because 
it does not align videos to a second modality (class-attribute vectors or semantic word embeddings).
Training VE proceeds by first sampling a set $\mathcal{Y}_{\text{batch}} \subset \mathcal{Y}_{\text{train}}$ of $n$ training classes. Then, a batch is formed by embedding $k$ instances of each class with a neural network, $f_\theta$:
\begin{equation}
\mathcal{B}_v = \{ (f_\theta(\xv_{ij}), y_{ij}) | y_{ij} \in \mathcal{Y}_{\text{batch}}\}_{i=1:n, \; j=1:k}.
\end{equation}
We estimate $\theta$ via backpropagation to minimize a deep metric-learning (DML) loss denoted by $\mathcal{L}_\text{DML}\big(\mathcal{B}_v\big)$. 

\paragraph{Word Embedding (WE)} To extend VE for CM-FSG, WE maps $\mathbf{x}^v$ directly to a word-embedding space of the class labels, denoted $b(\xl)$. 
$\mathcal{L}_\text{WE}$ combines $\mathcal{L}_\text{DML}$ with an alignment term between video and word embeddings:
\begin{equation}
\begin{split}
\label{loss:v2w}
&\mathcal{B}_{v,l} = \{ (f_\theta(\xv_{ij}), b(\xl_{ij})\}_{i=1:n, \; j=1:k}, \\
&\mathcal{L}_{\text{WE}} = \lambda \mathcal{L}_{\text{DML}}(\mathcal{B}_v) + \mathbb{E}_{\mathcal{B}_{v,l}} || f_\theta(\xv) - b(\xl) ||_2^2, 
\end{split}
\end{equation}
where $\mathcal{B}_v$ is defined as for VE. When $\lambda = 0$, this is equivalent to the loss from \cite{Socher2013}. 
When $\lambda > 0$,
this approach is similar to \cite{Hahn2019},
except (1)  we use $\mathcal{L}_{\text{DML}}(\mathcal{B}_v)$ instead of a linear layer 
trained with softmax cross-entropy and (2) 
we use mean-squared-error (MSE) between $f_\theta(\xv)$ and $b(\xl)$ 
instead of a contrastive loss. 

\paragraph{Joint Embedding (JE)} The downside of WE is that MSE imposes direct alignment between $f_\theta(\xv)$ and 
$b(\xl)$ (i.e., the model is encouraged to throw away visual features that are not represented in $b(\xl)$). Instead, JE 
maps both videos and word embeddings to a shared, joint 
embedding space. 
To do this, we train another neural network, $h_{\phi}$, that maps the word embeddings of the labels into a latent space of the same dimensionality as $f_\theta(\xv)$.
The embeddings are thus modality-agnostic, which lets us apply a cross-modal metric-learning loss to a shared batch defined as the union of the video and (twice-embedded) label batches: 

\vspace{-0.1in}
\begin{equation}
\begin{split}
&\mathcal{B}_h = \{ (h_\phi\big(b(\xl_{ij})\big), y_{ij}) | y_{ij} \in \mathcal{Y}_{\text{batch}}\}_{i=1:n, \; j=1}, \\
&\mathcal{L}_\text{JE} = \mathcal{L}_\text{DML}(\mathcal{B}_v \cup \mathcal{B}_h ), 
\end{split}
\end{equation}
where $\mathcal{B}_v$ is defined as for VE.

\section{Experiments and Conclusions}
\label{results}

\begin{table}[t!]
    \centering
    \small
    \input{split_stats_tyler}
    \caption{Counts of classes in each split, broken down by set (Train, Validation, Test) and by type (Held-out Noun and Held-out Verb).}
    \label{tab:split_stats_tyler}
\end{table}

Using the EPIC-KITCHENS training set, we constructed three new open-set splits, each with its own train, validation, and test set, where the classes are defined by the verb- and primary-noun-class as in the EPIC-KITCHENS challenge.
Within a split, classes are disjoint across train, validation, and test, as standard for the open-set setting.
We further sub-divided the test classes into (1) those with a held-out verb (HoV), but trained noun, 
(2) those with a held-out noun (HoN), but trained verb, and 
(3) those with a held-out verb and noun. 
Class-counts for each split are given in Table \ref{tab:split_stats_tyler}.
Since few classes fall into (3), we report performance on HoV, HoN, and the entire test set, denoted `All Test'.
Further details on the splits are provided in Appendix \ref{appendix:split_details}.

For all methods, $f_\theta$ is an I3D \cite{carreira2017quo} network, followed by an LSTM which collapses remaining timesteps into a single latent vector, the latent word-embedding, $b$, is a frozen, pretrained BERT model \cite{Wolf2019HuggingFacesTS}, 
and $h_\phi$ is a single fully-connected layer. For $\mathcal{L}_\text{DML}$ we experimented with both histogram loss \cite{Ustinova2016} and multi-similarity (multi-sim) loss \cite{Wang2019}, and all embeddings were L2-normalized (in the case of BERT, this was done post-hoc). These backbones are shared between the FSG and CM-FSG models in all of our experiments. 
We consider two variations of WE, one with $\lambda = 0$ (no metric-learning loss) which we denote \welambdazero~and a second \welambdaten, where $\lambda$ was chosen based on validation performance. To compute accuracy, we use a $\kappa$-nearest neighbor classifier over the embeddings, where $\kappa = k$. 

Figure \ref{fig:results} shows classification accuracy across the three methods for each split, test subset, and generalization task where $k = 1$ and $n = 5$.
We also explored FSG with $k \in \{1, 5\}$ and $n \in \{5, 20\}$, as well as CM-FSG with $k = 1$ and $n = 20$, and observed identical trends.
Figures and tables containing all results are provided in Appendix \ref{appendix:additional_results}.
Our unified framing of FSG and ZSG (as CM-FSG) allows us to compare performance of methods on both tasks for the first time. 
First, we observe that among CM-FSG-capable methods, the ones incorporate a metric-learning objective (\welambdaten~and JE) reliably 
outperform \welambdazero, a method designed for cross-modal prediction, on CM-FSG. 
Second, the joint embedding (JE) method leads to strictly superior CM-FSG and equivalent FSG when compared to VE and WE, indicating that among methods explored, there appears to be minimal trade-off between FSG and CM-FSG performance.
Third, we note that while there is some variability in performance across splits, it is smaller than variability across methods. 
This provides some evidence that our results are reliable, and that the
splitting procedure generates useful, novel evaluation splits of the EPIC-KITCHENS dataset. 
Incidentally, we find that the multi-sim loss systematically outperforms histogram loss, matching results from \cite{Ustinova2016,Wang2019}.

Our results, although preliminary, provide a strong baseline for comparison. 
We plan to exploit the broader framework defined here to further explore the space of evaluation paradigms for open-set classification. 
For example, the textual descriptions provided for action segments in EPIC-KITCHENS contain information beyond the verb and primary noun. These longer descriptions
could serve as a more informative input for cross-modal inference, and would enable evaluation of CM-FSG with $k > 1$.
We also plan to explore mixed-modal FSG, where the support sets contain a mixture of video and language samples.

{\small
\section*{Acknowledgments}
We thank for Ruta Desai, James Hillis, Kevin Carlberg, and Brenden Lake for helpful discussion and feedback.
}

{\small
\bibliographystyle{ieee_fullname}
\bibliography{egbib}
}

\input{appendix}

\end{document}

%% file: split_stats_tyler.tex
\begin{tabular}{c c c c c c c c}
\toprule
Split & Train & \multicolumn{3}{c}{Validation} & \multicolumn{3}{c}{Test} \\ 
\cmidrule(lr){3-5}\cmidrule(lr){6-8}
& & HoV & HoN & All & HoV & HoN & All \\
\midrule
1	&1715	&158	&102	&262	&248	&249	&536 \\
2	&1732	&135	&97	&239	&257	&247	&542 \\
3	&1731	&130	&104	&239	&280	&238	&543 \\
\midrule
\end{tabular}

%% file: appendix.tex
\clearpage

\appendix
\section{Experimental Details}
Both VE and JE are trained using a 256-dimensional embedding, while WE operates in the same space as the BERT embeddings, which has 768 dimensions.
The I3D backbone network is initialized using inflated ImageNet features \cite{carreira2017quo}.
The LSTM appended onto the I3D backbone network has a single layer with $d$ hidden units, where $d$ is the dimensionality of the embedding space, and is initialized using samples from a standard normal distribution.

\paragraph{Training Details}
For training and validation, we randomly sample two-second video clips within the labeled beginning and end frames, at 24 FPS, where each frame is resized to $256 \times 256$. 
During training, we augment the data by randomly applying a horizontal flip/mirror to all of the frames in each video clip.
For testing, we use the same procedure, but the clips are sampled to be the central 48 frames without mirroring.
For video clips less than two seconds, we add zero-padding.
The LSTM only processes non-padded frames.

To construct the batches used for training and validation, we sample $n=12$ classes and up to $k=8$ instances per class.
We ensure the batch contains at least 36 total instances or it is resampled.
For $\mathcal{B}_{v,l}$, we need a parallel set of word- and video-embeddings. 
Since there is only one word embedding per class, we create $k$ copies of it when constructing the batch.

In all experiments, models are fit with the Adam
optimizer. 
The initial learning rate is set to $1 \times 10^{-5}$, and is multiplied by 0.8 every 15,000 training batches.
Every 500 training steps, validation loss is averaged over 250 batches.
The model with the lowest validation loss is used for final evaluation.
Models are trained for a maximum of 75,000 batches, but could stop early based on a patience parameter that checks if the validation loss has decreased in the previous 15,000 batches.

\section{Split Details}
\label{appendix:split_details}

\begin{figure*}[ht]
    \centering
    \begin{minipage}[t]{0.45\textwidth}
        \centering
        \includegraphics[scale=0.35]{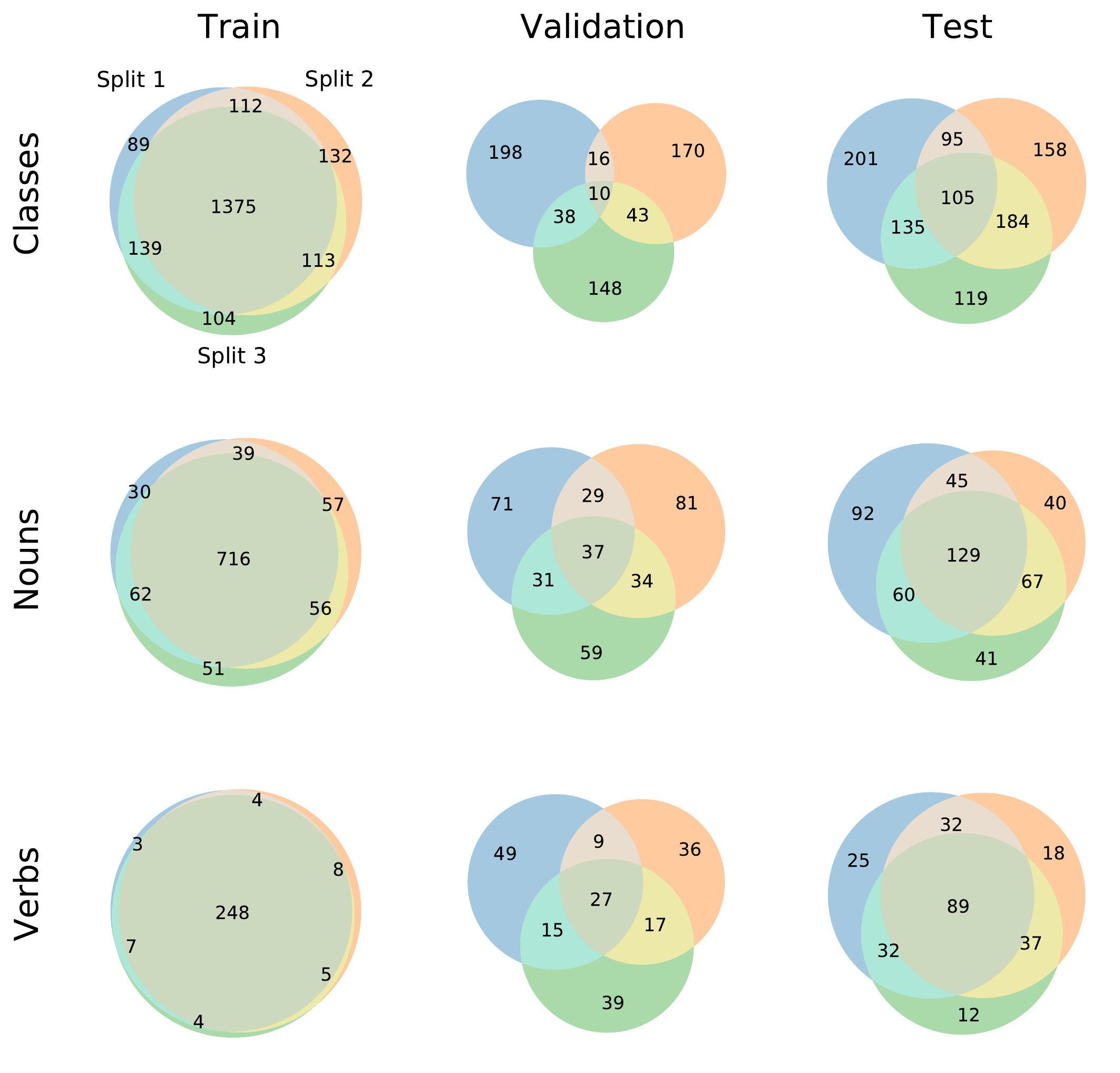}
        \caption{Venn diagrams showing overlap in splits for the train/validation/test sets, broken down by class, noun, and verb. 
        Each colored circle corresponds to one of the three splits. The overlapping regions between circles are annotated with the number of classes/nouns/verbs corresponding to that region.
        }
        \label{fig:split-overlap}
    \end{minipage}
    \hfill
    \begin{minipage}[t]{0.48\textwidth}
        \centering
        \includegraphics[scale=0.35]{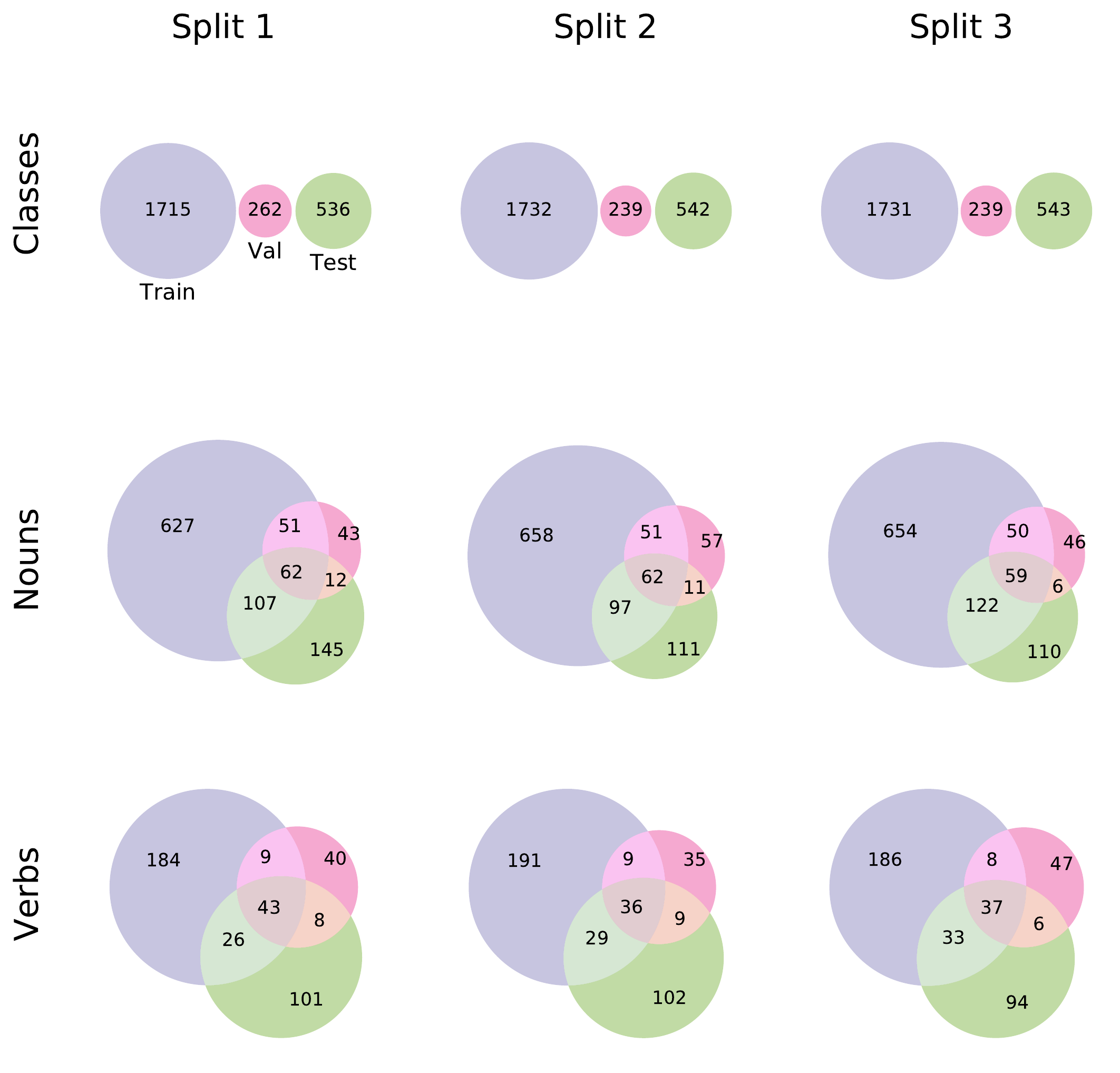}
        \caption{Venn diagrams showing overlap in train/validation/test sets for each split, broken down by class, noun, and verb. As expected in the open-set setting, the classes are all distinct between training, validation and test. At the same time, the validation and test splits include some nouns and verbs not seen at all during training (i.e.\ the HoN and HoV subsets), and some others that were seen as part of a different class context.}
        \label{fig:class-split-details}
    \end{minipage}
\end{figure*}

To generate the novel splits, we first cross-tabulated the verb and noun classes in the original training set, so that we could consider the dataset at the class rather than instance level. Next, we constructed a set of verbs and nouns eligible to be included in the validation and test sets by excluding verbs that appeared in fewer than $v_l$ contexts (i.e.\ with fewer than that many nouns) or those that appeared in more than $v_u$ contexts. We did the same for nouns with cutoffs $n_l$ and $n_u$. We did this to ensure there were sufficiently varied noun and verb contexts in the training set, and to ensure there were no singleton and near-singleton classes in the validation or test sets. Next, among the remaining classes we uniformly sampled $p_v$ verbs and $p_n$ nouns to be included in the validation/test sets, and further subsampled those into validation and test sets with proportions $p_v^t, p_n^t$. We selected all of the parameters $(v_l, v_y, n_l, n_u, p_v, p_n, p_v^t, p_n^t)$ by trial and error so that the number of classes in each held out subset (HoN validation, HoN test, HoV validation, HoV test) were roughly comparable. We next performed the same procedure for different seeds of the pseudo-random number generator, and retained splits where the counts were mostly balanced. Figure~\ref{fig:split-overlap} shows the number of overlapping classes, nouns, and verbs in each of our three novel splits. Note that while the training sets are fairly similar (owing to our class eligibility cutoff above), there is substantial variability in the classes, nouns, and verbs included in the validation and test sets between our splits, providing support for the success of our splitting procedure. This variability is likely contributing to the variability in results across splits. Figure~\ref{fig:class-split-details} shows the counts of overlapping classes, nouns, and verbs between the training, validation and test sets for each split. Consistent with the open-set setting, there are no overlapping classes between the sets, but there are some overlapping nouns and verbs, allowing us to evaluate performance for held-out nouns and verbs separately from overlapping classes. 

\section{All Results}
\label{appendix:additional_results}

\begin{figure*}[htbp]
\centering
\includegraphics[width=\textwidth]{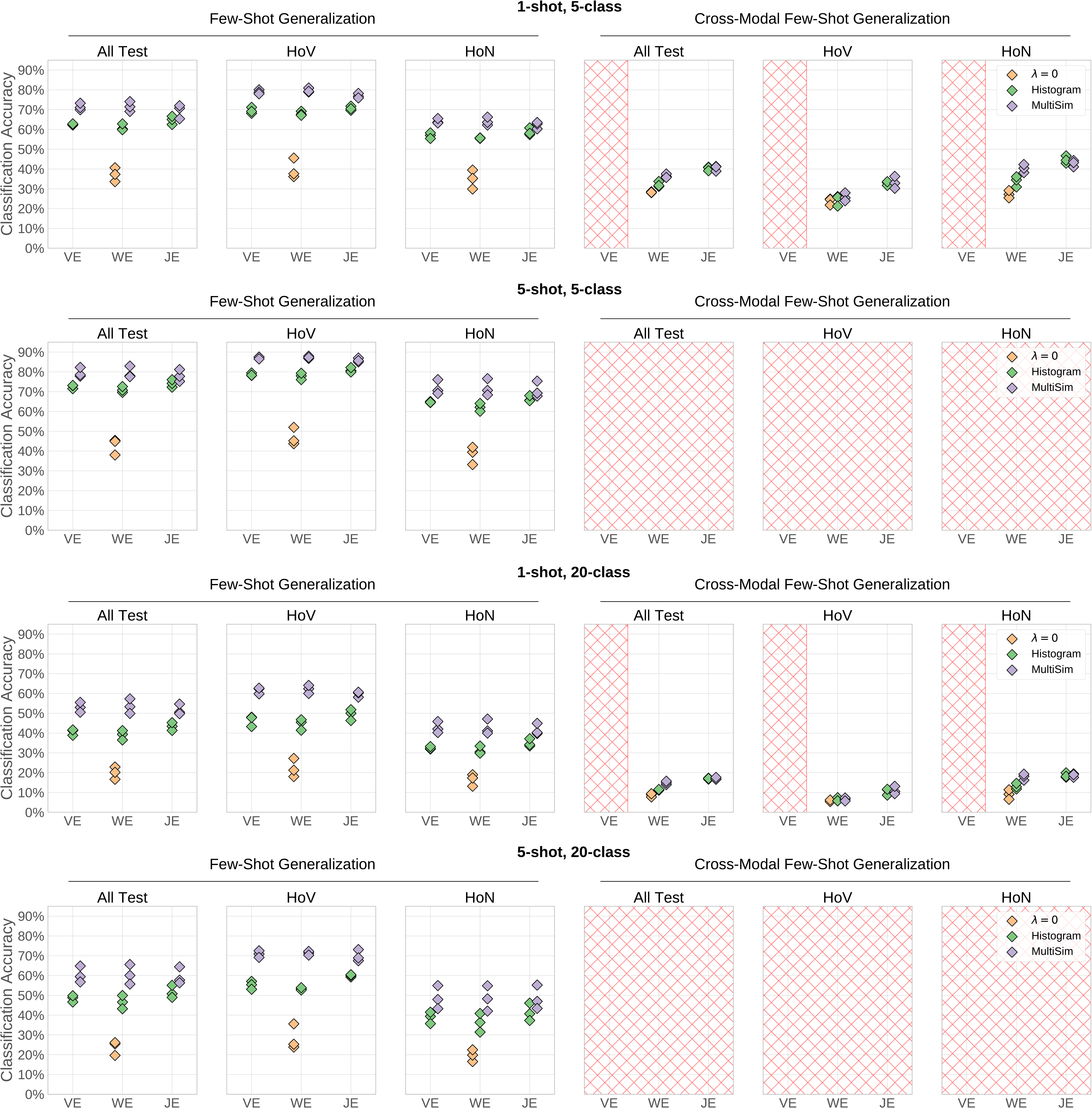}
\caption{Classification accuracy, computed over 500 test episodes, for the video embedding (VE), word embedding (WE), and joint embedding (JE). Each row in the plot corresponds to a setting of $k$ (`shot') and $n$ (`class'). $m = 20$ in all cases. Each pane is characterized according to a generalization task (FSG or CM-FSG) and a subset of the test set (All Test, HoV, or HoN). For a given generalization task, test subset, and method, the same-colored points represent performance on each of the three data splits. The red hatching indicates that the given method(s) could not be used to compute accuracy. For all settings, VE doesn't support CM-FSG. Furthermore, CM-FSG is only valid when $k = 1$, since we use class-labels as the support modality.}
\label{fig:all_results}
\end{figure*}

Results for FSG with $k \in \{1, 5\}$ and $n \in \{5, 20\}$, along with CM-FSG results with $k = 1$ and $n \in \{5, 20\}$ are presented in Figure \ref{fig:all_results}.
Tabulated results are included in Table \ref{table:fsg_results} for FSG and Table \ref{table:cm_fsg_results} for CM-FSG.

\begin{table*}[h!]
\caption{Tabulated FSG classification-accuracy results for the video embedding (VE), word embedding (WE), and joint embedding (JE). These results match those presented in Figure \ref{fig:all_results}. The three values grouped together in each row for a given class-type (All Test, HoV, HoN) correspond to the performance on splits 1, 2, and 3, respectively.}
\vspace{0.1in}
\label{table:fsg_results}
\definecolor{hilitecolor3}{RGB}{235,235,235}
\newcommand{\hiz}[1]{\colorbox{hilitecolor3}{\parbox{\widthof{#1}}{#1}}}
\begin{center}
\begin{small}

\begin{tabular}{cc | ccc | ccc | ccc}
\multicolumn{10}{l}{\hiz{FSG: 1-shot, 5-class}} \\
\toprule
& & \multicolumn{3}{c|}{All Test} & \multicolumn{3}{c|}{HoV} & \multicolumn{3}{c}{HoN} \\
\midrule
\multirow{2}{*}{VE} & Histogram & 62.2 & 62.6 & 62.8 & 71.2 & 68.2 & 69.1 & 57.0 & 58.2 & 55.4 \\
& MultiSim & 69.9 & 71.1 & 73.3 & 80.0 & 78.8 & 78.0 & 63.6 & 63.3 & 65.5 \\
\midrule
\multirow{3}{*}{WE} & $\lambda=0$ & 33.6 & 40.6 & 37.3 & 36.2 & 45.5 & 37.6 & 29.9 & 39.5 & 35.3 \\
& Histogram & 60.2 & 59.9 & 62.8 & 69.2 & 67.6 & 67.1 & 55.4 & 55.7 & 55.6 \\
& MultiSim & 69.1 & 71.4 & 74.1 & 80.9 & 78.9 & 79.2 & 62.2 & 63.6 & 66.2 \\
\midrule
\multirow{2}{*}{JE} & Histogram & 62.5 & 65.0 & 66.6 & 71.8 & 69.7 & 70.5 & 57.4 & 60.8 & 58.0 \\
& MultiSim & 65.4 & 70.8 & 72.0 & 76.7 & 78.2 & 75.9 & 60.3 & 62.8 & 63.5 \\
\end{tabular}

\vspace{0.1in}

\begin{tabular}{cc | ccc | ccc | ccc}
\multicolumn{10}{l}{\hiz{FSG: 5-shot, 5-class}} \\
\toprule
& & \multicolumn{3}{c|}{All Test} & \multicolumn{3}{c|}{HoV} & \multicolumn{3}{c}{HoN} \\
\midrule
\multirow{2}{*}{VE} & Histogram & 71.5 & 73.1 & 73.2 & 79.4 & 78.2 & 78.3 & 65.0 & 64.5 & 64.6 \\
& MultiSim & 77.8 & 78.5 & 82.3 & 87.5 & 87.5 & 86.6 & 70.6 & 69.1 & 76.1 \\
\midrule
\multirow{3}{*}{WE} & $\lambda=0$ & 38.0 & 45.4 & 44.9 & 43.7 & 52.0 & 45.2 & 33.2 & 39.4 & 41.9 \\
& Histogram & 69.6 & 70.7 & 72.6 & 78.0 & 76.1 & 79.3 & 62.2 & 60.1 & 64.1 \\
& MultiSim & 78.0 & 77.6 & 82.9 & 88.0 & 86.8 & 87.3 & 70.7 & 68.4 & 76.6 \\
\midrule
\multirow{2}{*}{JE} & Histogram & 72.3 & 74.2 & 76.0 & 80.6 & 80.0 & 82.2 & 65.6 & 65.4 & 68.0 \\
& MultiSim & 75.2 & 77.8 & 81.2 & 85.0 & 87.0 & 85.7 & 67.8 & 69.3 & 75.3 \\
\end{tabular}

\vspace{0.1in}

\begin{tabular}{cc | ccc | ccc | ccc}
\multicolumn{10}{l}{\hiz{FSG: 1-shot, 20-class}} \\
\toprule
& & \multicolumn{3}{c|}{All Test} & \multicolumn{3}{c|}{HoV} & \multicolumn{3}{c}{HoN} \\
\midrule
\multirow{2}{*}{VE} & Histogram & 41.3 & 38.9 & 41.6 & 48.0 & 43.3 & 47.8 & 31.9 & 32.2 & 33.2 \\
& MultiSim & 53.0 & 50.5 & 55.5 & 62.4 & 59.8 & 62.7 & 42.0 & 40.3 & 45.9 \\
\midrule
\multirow{3}{*}{WE} & $\lambda=0$ & 16.7 & 22.9 & 20.2 & 18.0 & 27.2 & 21.3 & 13.1 & 19.0 & 17.3 \\
& Histogram & 39.4 & 36.5 & 41.2 & 45.5 & 41.5 & 46.8 & 30.5 & 29.9 & 33.4 \\
& MultiSim & 53.3 & 50.0 & 57.3 & 62.4 & 60.0 & 64.1 & 41.1 & 40.0 & 47.1 \\
\midrule
\multirow{2}{*}{JE} & Histogram & 43.4 & 41.3 & 45.3 & 50.0 & 46.4 & 51.8 & 33.5 & 34.1 & 37.2 \\
& MultiSim & 50.5 & 49.8 & 54.7 & 58.1 & 60.3 & 60.7 & 39.7 & 40.3 & 44.9 \\
\end{tabular}

\vspace{0.1in}

\begin{tabular}{cc | ccc | ccc | ccc}
\multicolumn{10}{l}{\hiz{FSG: 5-shot, 20-class}} \\
\toprule
& & \multicolumn{3}{c|}{All Test} & \multicolumn{3}{c|}{HoV} & \multicolumn{3}{c}{HoN} \\
\midrule
\multirow{2}{*}{VE} & Histogram & 48.8 & 46.6 & 49.8 & 57.0 & 55.4 & 53.1 & 39.4 & 35.7 & 41.5 \\
& MultiSim & 59.4 & 56.8 & 64.8 & 70.9 & 72.5 & 69.2 & 47.9 & 43.3 & 54.9 \\
\midrule
\multirow{3}{*}{WE} & $\lambda=0$ & 19.6 & 25.5 & 26.1 & 23.9 & 35.6 & 25.3 & 16.5 & 19.8 & 22.5 \\
& Histogram & 46.6 & 43.2 & 49.9 & 53.2 & 52.7 & 53.9 & 36.3 & 31.4 & 40.8 \\
& MultiSim & 60.1 & 55.7 & 65.6 & 71.4 & 72.2 & 70.3 & 48.2 & 42.0 & 54.8 \\
\midrule
\multirow{2}{*}{JE} & Histogram & 50.8 & 49.0 & 55.1 & 59.3 & 60.0 & 60.4 & 40.7 & 37.3 & 46.1 \\
& MultiSim & 57.7 & 56.3 & 64.4 & 67.5 & 73.1 & 68.9 & 46.9 & 43.4 & 55.2 \\
\end{tabular}

\end{small}
\end{center}
\end{table*}

\begin{table*}[h!]
\caption{Tabulated CM-FSG classification-accuracy results for the video embedding (VE), word embedding (WE), and joint embedding (JE). These results match those presented in Figure \ref{fig:all_results}. The three values grouped together in each row for a given class-type (All Test, HoV, HoN) correspond to the performance on splits 1, 2, and 3, respectively.}
\vspace{0.1in}
\label{table:cm_fsg_results}
\definecolor{hilitecolor3}{RGB}{235,235,235}
\newcommand{\hiz}[1]{\colorbox{hilitecolor3}{\parbox{\widthof{#1}}{#1}}}
\begin{center}
\begin{small}

\begin{tabular}{cc | ccc | ccc | ccc}
\multicolumn{10}{l}{\hiz{CM-FSG: 1-shot, 5-class}} \\
\toprule
& & \multicolumn{3}{c|}{All Test} & \multicolumn{3}{c|}{HoV} & \multicolumn{3}{c}{HoN} \\
\midrule
\multirow{3}{*}{WE} & $\lambda=0$ & 28.1 & 28.5 & 28.2 & 24.8 & 24.6 & 21.8 & 27.2 & 25.4 & 29.2 \\
& Histogram & 31.2 & 33.7 & 31.7 & 26.1 & 25.7 & 21.3 & 30.9 & 34.5 & 36.1 \\
& MultiSim & 36.3 & 37.5 & 35.7 & 25.5 & 28.0 & 23.9 & 38.2 & 40.4 & 42.3 \\
\midrule
\multirow{2}{*}{JE} & Histogram & 40.9 & 40.7 & 39.1 & 33.2 & 31.7 & 33.6 & 43.0 & 46.7 & 44.5 \\
& MultiSim & 41.3 & 38.9 & 41.1 & 32.9 & 30.3 & 36.3 & 41.1 & 44.3 & 43.3 \\
\end{tabular}

\vspace{0.1in}

\begin{tabular}{cc | ccc | ccc | ccc}
\multicolumn{10}{l}{\hiz{CM-FSG: 1-shot, 20-class}} \\
\toprule
& & \multicolumn{3}{c|}{All Test} & \multicolumn{3}{c|}{HoV} & \multicolumn{3}{c}{HoN} \\
\midrule
\multirow{3}{*}{WE} & $\lambda=0$ & 9.0 & 7.7 & 9.4 & 6.2 & 5.4 & 6.2 & 9.1 & 6.5 & 11.4 \\
& Histogram & 11.1 & 11.4 & 11.5 & 7.4 & 5.8 & 5.8 & 11.8 & 12.8 & 14.6 \\
& MultiSim & 14.0 & 14.9 & 15.7 & 6.5 & 7.2 & 5.7 & 16.2 & 18.2 & 19.3 \\
\midrule
\multirow{2}{*}{JE} & Histogram & 16.7 & 17.2 & 17.1 & 11.4 & 8.7 & 11.6 & 17.7 & 19.9 & 18.1 \\
& MultiSim & 16.7 & 16.8 & 17.6 & 10.5 & 9.4 & 13.1 & 17.7 & 19.4 & 19.0 \\
\end{tabular}

\end{small}
\end{center}
\end{table*}